# A Comprehensive Trainable Error Model for Sung Music Queries

**Colin J. Meek**                                        CMeek@microsoft.com
*Microsoft Corporation, SQL Server,*
*Building 35/2165, 1 Microsoft Way, Redmond WA 98052 USA*

**William P. Birmingham**                           WPBirmingham@gcc.edu
*Grove City College, Math and Computer Science,*
*Faculty Box 2655, 100 Campus Drive, Grove City PA 16127 USA*

## Abstract

We propose a model for errors in sung queries, a variant of the hidden Markov model (HMM). This is a solution to the problem of identifying the degree of similarity between a (typically error-laden) sung *query* and a potential *target* in a database of musical works, an important problem in the field of music information retrieval. Similarity metrics are a critical component of "query-by-humming" (QBH) applications which search audio and multimedia databases for strong matches to oral queries. Our model comprehensively expresses the types of *error* or variation between target and query: cumulative and non-cumulative local errors, transposition, tempo and tempo changes, insertions, deletions and modulation. The model is not only expressive, but automatically trainable, or able to learn and generalize from query examples. We present results of simulations, designed to assess the discriminatory potential of the model, and tests with real sung queries, to demonstrate relevance to real-world applications.

## 1. Introduction

Many approaches have been proposed for the identification of viable targets for a query in a music database. Query-by-humming systems attempt to address the needs of the non-expert user, for whom a natural query format – for the purposes of finding a tune, hook or melody of unknown providence – is to sing it. Our goal is to demonstrate a unifying model, expressive enough to account for the complete range of modifications observed in the performance and transcription of sung musical queries. Given a complete model for singer error, we can accurately determine the likelihood that, given a particular *target* (or song in a database), the singer would produce some *query*. These likelihoods offer a useful measure of *similarity*, allowing a query-by-humming (QBH) system to identify strong matches to return to the user.

Given the rate at which new musical works are recorded, and given the size of multimedia databases currently deployed, it is generally not feasible to learn a separate model for each target in a multimedia database. Similarly, it may not be possible to customize an error model for every individual user. As such, a QBH matcher must perform robustly across a broad range of songs and singers. We develop a method for training our error model that functions well across singers with a broad range of abilities, and successfully generalizes to works for which no training examples have been given (see Section 11). Our approach (described in Section 9) is an extension of a standard re-estimation algorithm (Baum &

Eagon, 1970), and a special case of Expectation Maximization (EM) (Dempster, Laird, & Jain, 1977). It is applicable to hidden Markov models (HMM) with the same dependency structure, and is demonstrated to be convergent (see Appendix A).

## 2. Problem Formulation and Notation

An assumption of our work is that pitch and IOI adequately represent both the target and the query. This limits our approach to monophonic lines, or sequences of non-overlapping note events. An event consists of a $\langle Pitch, IOI \rangle$ duple. *IOI* is the time difference between the onsets of successive notes, and *pitch* is the MIDI note number[1].

We take as input a *note-level* abstraction of music. Other systems act on lower-level representations of the query. For instance, a frame-based frequency representation is often used (Durey, 2001; Mazzoni, 2001). Various methods for the translation of frequency and amplitude data into note abstraction exist (Pollastri, 2001; Shifrin, Pardo, Meek, & Birmingham, 2002). Our group currently uses a transcriber based on the Praat pitch-tracker (Boersma, 1993), designed to analyze voice pitch contour. A sample Praat analysis is shown in Figure 1. In addition to pitch extraction, the query needs to be *segmented*, or organized into contiguous events (notes). The query transcription process is described in greater detail in Section 3. Note that these processes are not perfect, and it is likely that error will be introduced in the transcription of the query.

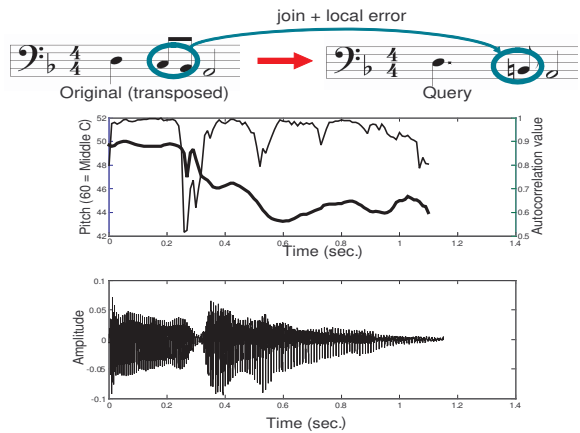

Figure 1: Sample query transcription: from *Hey Jude* by the Beatles, the word "better"

Restricting ourselves to this event description of target and query ignores several elements of musical style, including dynamics, articulation and timbre, among others. Objectively and consistently characterizing these features is quite difficult, and as such we have little confidence they can be usefully exploited for music retrieval at this point. We acknowledge, however, the importance of such elements in music query/retrieval systems in general. They will likely prove essential in refining or filtering the search space (Birmingham, Pardo,

---

1. Musical Instrument Digital Interface (MIDI) has become a standard electronic transmission and storage protocol/format for music. MIDI note numbers essentially correspond to the keys of a piano, where 'middle C' corresponds to the integer value 60.

Meek, & Shifrin, 2002; Birmingham, Dannenberg, Wakefield, Bartsch, Bykowski, Mazzoni, Meek, Mellody, & Rand, 2001).

We further simplify the representation by using IOI quantization and by representing pitch in terms of pitch class. IOI is quantized to a logarithmic scale, using $q = 29$ quantization levels, within the range 30 msec. to 3840 msec., chosen such that there are precisely four gradations between an eighth note and sixteenth note (or quarter note and sixteenth note, and so forth.) This representation mirrors conventional notation in Western music, in which the alphabet of rhythmic symbols (eighth, quarter, half, etc.) corresponds to a logarithmic scale on duration (see Figure 2), and has been shown not to substantially influence discrimination between potential targets in a database (Pardo & Birmingham, 2002).

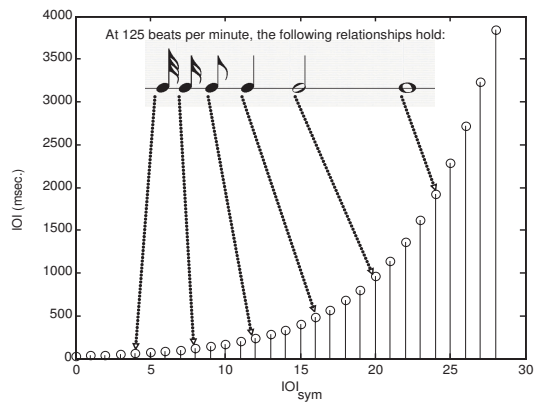

Figure 2: IOI quantization

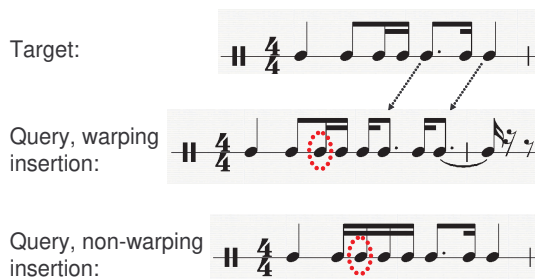

Figure 3: Warping and non-warping insertions.

Pitch is translated to *pitch class*, a representation where all notes are projected into a single octave, and are considered in the context of the 12-tone, well-tempered scale. For instance, the frequency 453 Hz is "binned" into MIDI note number 70. The corresponding pitch class is $\mod_{12}(70) = 10$. This addresses two issues: octave errors are quite common in some transcriber systems, and pitch class is an effective, if imperfect, musical (Pardo, 2002) and perceptual (Bartsch & Wakefield, 2001) abstraction. In addition, pitch class substantially reduces the model's "alphabet" size.

In our implementation, pitch bins are not fixed, but vary from query to query. We first convert frequency ($f$) to what might be termed a floating-point MIDI note representation ($m$), using the assumption of equal-temperament (twelve equally spaced semitones per octave) and, according to convention, translating a frequency of 440 Hz to 'A' above middle 'C', or MIDI note value 69: $m = 69 + 12 \log_2 \frac{f}{440}$. When we round to the nearest note number, we introduce a rounding error. To minimize this error, we add an offset to the note numbers, to account for any overall pitch tendency. For instance, consider a query consisting of a sequence of note numbers {48.4, 46.6, 44.4, 43.6}. This query tends to be sharp (or flat) by roughly a quarter-tone. By introducing an offset of +0.5 (a quarter-tone), we minimize the rounding error, and can thus more closely preserve the contour: without the offset, we round to {48, 47, 44, 44}; with the offset, we round to {49, 47, 45, 44}. Since transpositional invariance is a central feature of our model, the direction of the offset is irrelevant in this example. Adopting an approach proposed for a QBH "audio front end" (Pollastri, 2001), we consider several offsets ($O = \{0.0, 0.1, \ldots, 0.9\}$). Given a sequence of note numbers ($M = \{m1, m2, \ldots, m_n\}$), we choose the offset ($o \in O$) such that the mean error squared ($e = \frac{\sum_{i=1}^{n}[m+o-\mathrm{round}(m+o)]^2}{n}$) is minimized, and set $Pitch[i]$ equal to round($m_i + o$).

We choose discrete sets of symbols to represent pitch and duration since, as will be seen, a continuous representation would necessitate an unbounded number of states in our model. This second event representation is notated:

$$o_t = \langle P[t], R[t] \rangle \tag{1}$$

for queries (using the mnemonic shorthand $\mathbf{o}bservation = \langle \mathbf{P}itch, \mathbf{R}hythm \rangle$). Target events are similarly notated:

$$d_i = \langle P[i], R[i] \rangle \tag{2}$$

For clarity, we will return to the earlier representation $\langle Pitch[i], IOI[i] \rangle$ where appropriate. The second representation is derived from the first as follows, where 30 and 3840 are the IOI values associated with the centers of the shortest and longest bins, and $q$ is the number of IOI quantization bins:

$$P[i] = \mod_{12}(Pitch[i]) \tag{3}$$

$$R[i] = \mathrm{round}\left(\frac{\log IOI[i] - \log 30}{\log 3840 - \log 30} \cdot (q-1)\right) \tag{4}$$

The goal of this paper is to present a model for query errors within the scope of this simple event representation. We will first outline the relevant error classes, and then present an extended Hidden Markov Model accounting for these errors. Taking advantage of certain assumptions about the data, we can then efficiently calculate the likelihood of a target model generating a query. This provides a means of ranking potential targets in a database (denoted $\{D_1, D_2, \ldots\}$, where $D_i = \{d_1, d_2, \ldots\}$) given a query (denoted $O = \{o_1, o_2, \ldots\}$) based on the likelihood the models derived from those targets generated a given query. A summary of the notation used in this paper is provided in Appendix B.

## 3. Query Transcription

Translating from an audio query to a sequence of note events is a non-trivial problem. We now outline the two primary steps in this translation: frequency analysis and segmentation.

### 3.1 Frequency Analysis

We use the Praat pitch-tracker (Boersma, 1993), an enhanced auto-correlation algorithm developed for speech analysis, for this stage. This algorithm identifies multiple auto-correlation peaks for each analysis frame, and chooses a path through these peaks that avoids pitch jumps and favors high correlation peaks. For a particular frame, no peak need be chosen, resulting in gaps in the frequency analysis. In addition, the algorithm returns the auto-correlation value at the chosen peak (which we use as a measure of pitch-tracker confidence), and the RMS amplitude by frame (see Figure 1 for instance.)

## 3.2 Segmentation

A binary classifier is used to decide whether or not each analysis frame contains the beginning of a new note. The features considered by the classifier are derived from the pitch-tracker output. This component is currently in development at the University of Michigan. In its present implementation, a five-input, single-layer neural network performs the classification. We assign a single pitch to each note segment, based on the weighted average pitch by confidence of the frames contained in the segment. An alternative implementation is currently being explored, which treats the query analysis as a signal (ideal query) with noise, and attempts to uncover the underlying signal using Kalman-filter techniques.

## 4. Error Classes

A query model should be capable of expressing the following musical – or un-musical you might argue – transformations, relative to a target:

1. Insertions and deletions: adding or removing notes from the target, respectively. These *edits* are frequently introduced by transcription tools as well.

2. Transposition: the query may be sung in a different *key* or *register* than the target. Essentially, the query might sound "higher" or "lower" than the target.

3. Tempo: the query may be slower or faster than the target.

4. Modulation: over the course of a query, the transposition may change.

5. Tempo change: the singer may speed up or slow down during a query.

6. Non-cumulative local error: the singer might sing a note off-pitch or with poor rhythm.

## 4.1 Edit Errors

Insertions and deletions in music tend to influence surrounding events. For instance, when an insertion is made, the inserted event and its neighbor tend to occupy the temporal space of the original note: if an insertion is made and the duration of the neighbors is not modified, the underlying rhythmic structure (the beat) is changed. We denote this type of insertion a "warping" insertion. For instance, notice the alignment of notes after the warping insertion in Figure 3, indicated by the dotted arrows. The inserted notes are circled. For the non-warping insertion, the length of the second note is shortened to accommodate the new note.

   With respect to pitch, insertions and deletions do not generally influence the surrounding events. However, previous work assumes this kind of influence: noting that intervallic contour tends to be the strongest component in our memory of pitch; one researcher has proposed that insertions and deletions could in some cases have a "modulating" effect (Lemstrom, 2000), where the edit introduces a pitch offset, so that pitch intervals rather than the pitches themselves are maintained. We argue that *relative* pitch, with respect to the query as a whole, should be preserved. Consider the examples in Figure 4. The first row of numbers below the staff indicates MIDI note numbers, the second row indicates the intervals in semitones ('u' = up, 'd' = down.) Notice that the intervallic representation is preserved in the modulating insertion, while the overall "profile" (and key) of the line is maintained in the non-modulating insertion.

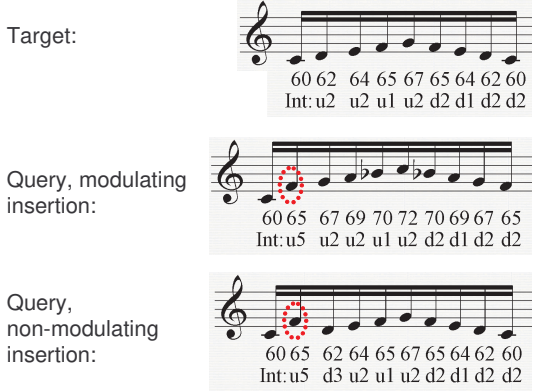

Figure 4: Modulating and non-modulating insertions

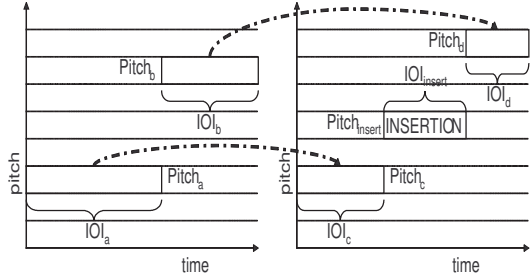

Figure 5: Insertion of a note event in a query

The effects of these various kinds of insertions and deletions are now formalized, with respect to a target consisting of two events $\{\langle Pitch_a, IOI_a\rangle, \langle Pitch_b, IOI_b\rangle\}$, and a query $\{\langle Pitch_c, IOI_c\rangle, \langle Pitch_{insert} IOI_{insert}\rangle \langle Pitch_d, IOI_d\rangle\}$, where $\langle Pitch_{insert} IOI_{insert}\rangle$ is the inserted event (see Figure 5). Note that deletion is simply the symmetric operation, so we show examples of insertions only:

- Effects of a warping insertion on IOI: $IOI_c = IOI_a$, $IOI_d = IOI_b$

- Effects of a non-warping insertion on IOI: $IOI_c = IOI_a - IOI_{insert}$, $IOI_d = IOI_b$

- Effects of a modulating insertion on pitch: $Pitch_c = Pitch_a$, $Pitch_d = Pitch_{insert} + \underbrace{Pitch_b - Pitch_a}_{\text{pitch contour}}$

- Effects of a non-modulating insertion on pitch: $Pitch_c = Pitch_a$, $Pitch_d = Pitch_b$

In our model, we deal only with non-modulating and non-warping insertions and deletions explicitly, based on the straightforward musical intuition that insertions and deletions tend to operate within a rhythmic and modal context. The other types of edit are represented *in combination with* other error classes. For instance, a modulating insertion is simply an insertion combined with a modulation.

Another motivation for our "musical" definition of edit is transcriber error. In this context, we clearly would not expect the onset times or pitches of surrounding events to be influenced by a "false hit" insertion or a missed note. The relationships amongst successive events must therefore be modified to avoid warping and modulation. Reflecting this bias, we use the terms "join" and "elaboration" to refer to deletions and insertions, respectively. A system recognizing musical variation (Mongeau & Sankoff, 1990) uses a similar notion of insertion and deletion, described as "fragmentation" and "consolidation" respectively.

## 4.2 Transposition and Tempo

We account for the phenomenon of persons reproducing the same "tune" at different speeds and in different registers or keys. Few people have the ability to remember and reproduce exact pitches (Terhardt & Ward, 1982), an ability known as "absolute" or "perfect" pitch. As such, transpositional invariance is a desirable feature of any query/retrieval model. The effect of transposition is simply to add a certain value to all pitches. Consider for example the transposition illustrated in Figure 6, Section a, of $Trans = +4$.

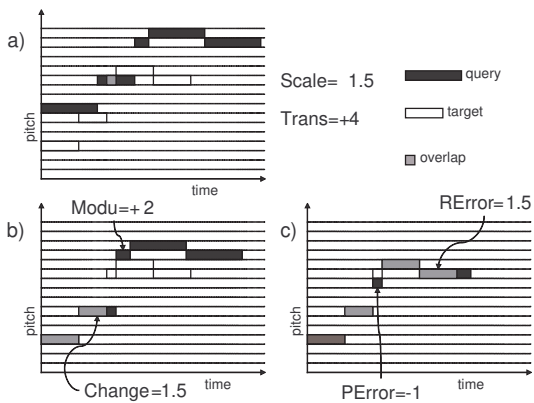

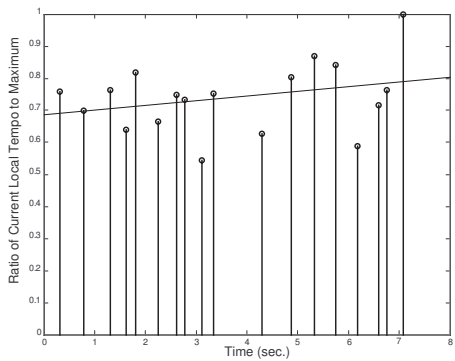

Figure 6: Error class examples, opening notes of Brahms' *Cello Sonata in e-minor*

Figure 7: Tempo increase

Tempo in this context is simply the translation of rhythm, which describes duration relationships, into actual time durations. Again, it is difficult to remember and reproduce an exact tempo. Moreover, it is very unlikely that two persons would choose the same metronome marking, much less unconstrained beat timing, for any piece of music. This is a natural "musical" interpretation. We measure tempo relative to the original using a scaling factor on rhythmic duration. Thus, if the query is 50% slower than the target, we have a scaling value of $Scale = 1.5$, as shown in Figure 6, Section a.

In practice, we use quantized tempo scaling and duration values. Note that addition in the logarithmic scale is equivalent to multiplication: with quantized IOI values, we replace floating point multiplication with integer addition when applying a scaling value. For instance, given our quantization bins, a doubling of tempo always corresponds to an addition of four: $Scale = 2.0 \leftrightarrow Scale_{quantized} = +4$.

## 4.3 Modulation and Tempo Change

Throughout a query, the degree of transposition or tempo scaling can change, referred to as *modulation* and *tempo change*, respectively. Consider a query beginning with the identity transposition $Trans = 0$ and identity tempo scaling $Scale = 1$, as in Figure 6, Section b. When a modulation or tempo change is introduced, it is always with respect to the previous transposition and tempo. For instance, on the third note of the example, a modulation of $Modu = +2$ occurs. For the remainder of the query, the transposition is equal to $0 + 2 = +2$, from the starting reference transposition of 0. Similarly, the tempo change of $Change = 1.5$ on the second note means that all subsequent events occur with a rhythmic scaling of $1 \cdot 1.5 = 1.5$.

Consider Figure 7, which plots the apparent tempo scaling in a rendition of "Row, Row, Row Your Boat" on a note-by-note basis. While our model considers several interpretations of such a rendition, one approach would be to consider a constantly increasing tempo, represented by the least-square deviation regression line, with local rhythmic errors (see Section 4.4), represented by the note-wise deviations from that line.

### 4.4 Local Pitch and IOI Errors

In addition to the "gross" errors we have discussed thus far, there are frequently local errors in pitch and rhythm. These errors are relative to the modifications described above. A local pitch error of $\Delta^{(P)}$ simply adds some value to the "ideal" pitch, where the ideal is determined by the relevant target note and the current transposition. A local IOI error of $\Delta^{(R)}$ has a scalar effect (or again, additive in the quantized domain) on the ideal IOI, derived from the relevant target note and the current tempo. Figure 6, Section c, shows examples of each error. Note that these errors do not propagate to subsequent events, and as such are termed *non-cumulative* or *local* errors. Transposition and tempo change are examples of *cumulative* error.

In some cases, there are multiple interpretations for the source of error in a query. Consider for instance Figure 8, which shows a specific interpretation of three disagreements between a target and query. The second note in the query is treated as a local pitch error of -1. The final two notes, which are a semi-tone sharper than expected (+1), are explained as a modulation. The error model, described in the next section, considers all possible interpretations, for instance considering the possibility that the error in the second note is accounted for by two modulations (before and after), and the final two notes by a pair of local errors. Depending on our expectation that such things might occur, one or the other interpretation might appear more likely. In general, we would prefer to find the most direct possible explanations for queries, since an increased likelihood of error in the model can be shown to reduce discrimination (see Section 10).

## 5. Existing Error Models

For edits, we assume that overall rhythm is maintained, and make the natural musical assumption that edits have a local impact on pitch. Many QBH applications adopt this approach to rhythm (Mazzoni, 2001; Meek & Birmingham, 2002; Pauws, 2002; McNab, Smith, Bainbridge, & Witten, 1997; McNab, Smith, Witten, Henderson, & Cunningham, 1996).

In this study, we are concerned primarily with local and cumulative error. Far less is known about this area. This is largely a matter of convenience: a particular musical representation will tend to favor one approach over the other. For instance, a pitch- and tempo-invariant representation (pitch interval and inter-onset interval ratio) (Shifrin et al., 2002; Pauws, 2002) establishes a new transposition and tempo context for each note, thus introducing the implicit assumption that all errors are cumulative (Pardo & Birmingham, 2002). A study of sung queries (Pollastri, 2001) determined that cumulative error is in fact far less common than local error, a conclusion supported by our studies.

Another approach to the differences in transposition and tempo context is to attempt multiple passes over a fixed context model, and evaluate error rigidly within each pass by comparing the query to various permutations of the target. Dynamic time-warping approaches (Mazzoni, 2001) and non-distributed HMM techniques (Durey, 2001) are well-suited to this technique. It is not possible to model, for instance, a modulation, using these methods, only local error. A preliminary QBH proposal (Wiggins, Lemstrom, & Meredith, 2002) recommends a similar approach, grouping together "transposition vectors" connecting query and target notes. Such approaches are amenable to extensions supporting cumulative error as well, but have not – to our knowledge – been extended in this way.

An alternative is to normalize the tempo of the query by either automated beat-tracking, a difficult problem for short queries, or, more effectively, by giving the querier an audible beat to sing along with – a simple enough requirement for users with some musical background (Chai, 2001). Again, there is an assumption that the transposition will not change during a query, but the beat-tracker can adapt to changing tempi.

## 5.1 Alternative Approaches

We are concerned primarily with sequence based approaches to music retrieval. It is possible to relax this assumption somewhat, by translating targets into Markov models where the state is simply a characteristic relationship between consecutive notes, allowing for loops in the model (Shifrin et al., 2002). Borrowing from the text search world, we can also model music as a collection of note $n$-grams, and apply standard text retrieval algorithms (Downie, 1999; Tseng, 1999). In query-by-humming systems, the user is searching for a song that "sounds like..." rather than a song that is "about" some short snippet of notes, if it makes sense to discuss music in these terms at all[2]. For this reason, we believe that sequence-based methods can more accurately represent music in this context.

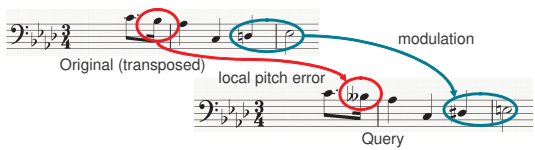

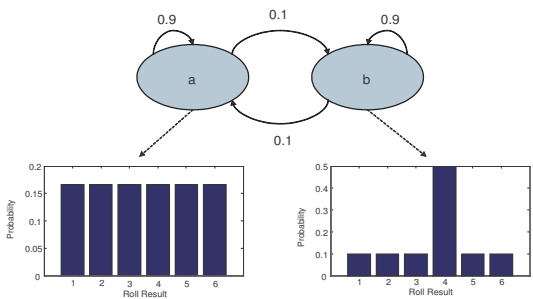

Figure 8: Portion of a query on the *American National Anthem*, including examples of modulation and local pitch error

Figure 9: Simple HMM, the dishonest gambler

---

2. Beethoven's Fifth Symphony is a notable exception

## 5.2 Indexing and Optimization Techniques

It should be noted that as a general model, JCS cannot take advantage of optimizations specific to certain specializations. For instance, the cumulative-only version is amenable to a *relative note representation*, a transposition- and tempo- invariant representation, obviating the need to compare a query with multiple permutations of the target.

Existing indexing techniques for string-edit distance metrics – for instance using suffix trees (Chávez & Navarro, 2002; Gusfield, 1997) and so-called 'wavelet' approximations (Kahveci & Singh, 2001) – are appropriate for $k$-distance searches, and thus might prove useful as a pre-filtering mechanism. However, they are not readily adaptable to more sophisticated probabilistic matching metrics, and, more importantly, rely on a *global alignment* assumption infeasible for QBH applications. A recent application of dynamic time-warping indexing to music retrieval (Zhu & Shasha, 2003) relies on the assumption that the query is globally aligned to the target, which is feasible only when the query covers precisely the material in the target.

Other indexing techniques can provide benefits even in the general case. Lexical trees are commonly used in speech recognition for large-vocabulary recognition (Huang, Acero, & Hon, 2001), and can be adapted to the alignment and metric requirements of JCS using similar trie structures. Linear search using sub-optimal heuristics has been applied to sequence matching in bio-informatics (Pearson, 1998; Altschul, Madden, Schaffer, Zhang, Zhang, Miller, & Lipman, 1997). These approaches are fast for massive databases, and reasonably accurate even with low similarity distance metrics. Optimal alignment techniques – adaptable to arbitrary probabilistic sequence alignment models – are also being developed using trie indices (Meek, Patel, & Kasetty, 2003) and parallel architectures (Shpaer, Robinson, Yee, Candlin, Mines, & Hunkapiller, 1996).

We are concerned primarily with determining appropriate metrics for QBH applications. Once this has been accomplished, it will be possible to choose and develop appropriate optimizations and indexing techniques for large-scale deployment.

## 6. Hidden Markov Models

Hidden Markov Models (HMM) are the basis for our approach. We will begin by describing a simple HMM, and then describe the extensions to the model necessary for the current task. As suggested by the name, HMMs contain hidden, or unobserved, states. As a simple example, consider a dishonest gambler, who is known to occasionally swap a fair dice for a loaded dice (with thanks to "Biological Sequence Analysis" (Durbin, Eddy, Krogh, & G.Mitchison, 1998) for the example). Unfortunately, it is impossible to observe (directly) which of the dice is being used, since they are visually indistinguishable. For this reason, we define two hidden states, $a$ and $b$, representing the conjectures that the gambler is using fair and loaded dice, respectively. Further, we represent our expectation that the gambler will switch dice or stay with a dice using a transition diagram, where the transitions have associated probabilities (see Figure 9). For instance, the arc from $a \rightarrow b$ is labelled 0.1, indicating that the probability of the gambler switching from the fair dice to the loaded dice after a roll is 0.1, or formally $Pr(q_{t+1} = b | q_t = a, \lambda)$ where $q_t$ is the current state at time interval $t$, and $\lambda$ is the model. What we *can* directly observe in this example is the result of each roll. While we do not know which dice is being used, we know some distribution over the roll values for each dice (shown at the bottom of Figure 9). These are refereed to as *observation* or *emission* probability functions, since they describe the probability of emitting a particular observation in a state.

In the music information-retrieval (MIR) context, we have a series of models, each representing a possible database target, and wish to determine which is most likely given a query, represented by a sequence of pitch and rhythm observations. To draw a parallel to our gambler example, we might want to determine whether we are dealing with the dishonest gambler described above, or an honest gambler (see Figure 10) who uses only fair dice. Given some sequence of observations, or dice rolls, we can determine the likelihood that each of the models generated that sequence.

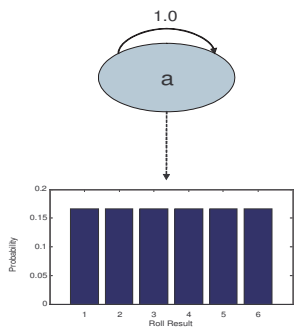

Figure 10: Simple HMM, the honest gambler

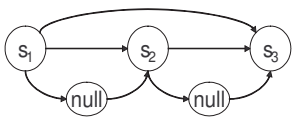

Figure 11: A model with skips and null states

## 6.1 Honest or Dishonest? An example

The strength of the HMM approach is that it is straightforward to determine the likelihood of any observation sequence if the transition probabilities and emission probabilities are known. Conceptually, the idea is to consider all possible paths through the model consistent with the observation sequence (e.g., the observed dice rolls), and take the sum of the probabilities of each path given those observations. For instance, the roll sequence {1, 5, 4} could be generated by one of four paths in the dishonest gambler model, assuming that the dishonest gambler always begins with the fair dice: $\{\{a, a, a\}, \{a, a, b\}, \{a, b, a\}, \{a, b, b\}\}$. The probability of the second path, for instance, is equal to the probability of the transitions ($Pr(a \rightarrow a) \cdot Pr(a \rightarrow b)$=$0.9 \cdot 0.1 = 0.09$) multiplied by the probabilities of the observations given the states in the path (the probability of rolling 1 then 5 with the fair dice, by the probability of rolling 4 with the loaded dice: $0.167 \cdot 0.167 \cdot 0.5 = 0.0139$), which is equal to 1.25e-3. To determine the likelihood of the observation sequence given the model, we simply take the sum of the probabilities of each path (3.75e-3 + 1.25e-3 + 2.78e-5 + 7.50e-4 = 5.78e-3.) The honest gambler is in effect a fully observable model, since there is only a single hidden state. Only one path through this model is possible, and its likelihood is therefore a function of the observation probabilities only since the path is deterministic (1.0 transition probabilities): $(0.167)^3 = 4.63e-3$. From this, we can conclude that the sequence of rolls 1, 5, 4 is more likely to have been generated by a dishonest gambler, though we should note that three rolls do not provide much evidence one way or the other!

## 7. Extended HMM

In the context of our query error model, we account for edit errors (insertions and deletions) in the "hidden" portion of the model. Using the notion of state "clusters," we account for transposition, modulation, tempo and tempo changes. Fine pitch and rhythm errors are accounted for in the observation distribution function.

### 7.1 State Definition

The state definition incorporates three elements: edit type ($E$dit), transposition ($K$ey) and tempo ($S$peed). States are notated as follows:

$$s_x = \langle E[x], K[x], S[x] \rangle, 1 \leq x \leq n \tag{5}$$

If $\mathbf{E}$ is the set of all edit types, $\mathbf{K}$ is the set of all transpositions, and $\mathbf{S}$ is the set of all tempi, then the set of all states $\mathbf{S}$ is equal to:

$$\mathbf{E} \times \mathbf{K} \times \mathbf{S} \tag{6}$$

We now define each of these sets:

#### 7.1.1 EDIT TYPE

For the sake of notational clarity, we do not enumerate the edit types in $\mathbf{E}$, but define them in terms of symbols that *indirectly refer* to events in the target sequence, encoding *position* information. There are three types of symbol:

- $Same_i$: refers to the correspondence between the $i^{\text{th}}$ note in the target and an event in the query.

- $Join_i^l$: refers to a "join" of $l$ notes, starting from the $i^{\text{th}}$ note in the target. In other words, a single note in the query replaces $l$ notes in the target.

- $Elab_{i,j}^m$: refers to the $j^{\text{th}}$ query note elaborating the $i^{\text{th}}$ target note. In other words, a single note in the target is replaced by $m$ notes in the query.

Notice that $Same_i = Join_i^1 = Elab_{i,1}^1$, each referring to a one-to-one correspondence between target and query notes. In our implementation, $Join_i^1$ plays all three roles. We generate a set of states for a given target consisting of, for *each* event in the target $d_i$:

- A $Same_i$ state.

- Join states $Join_i^l$, for $2 \leq l \leq L$ where $L$ is some arbitrary limit on the number of events that can be joined.

- Elaboration states $Elab_{i,j}^m$ for $2 \leq m \leq M$ and $1 \leq j \leq m$, where $M$ is some arbitrary limit on the length of elaborations.

Why do we have so many states to describe each event in the target? We wish to establish a one-to-one correspondence between hidden states and query events, to simplify the implementation, which is why we introduce multiple states for each elaboration. We choose not to implement joins by "skips" through a reduced set of states, or elaborations as null states, since as discussed, edits influence our interpretation of the underlying target events. Figure 11 illustrates a model with skips and null states. Given our definition of insertion and deletion, state $s_1$ would need separate emission probability tables for each outward arc (and thus would be functionally and computationally equivalent to the model we propose).

As mentioned, we explicitly handle only non-modulating and non-warping insertions and deletions (see Section 4.1). As such, when comparing target and query events with respect to a join, we generate a longer target note, with the sum duration of the relevant target events, and the pitch of the first. Similarly, for an elaboration, we treat a sequence of query notes as a single, longer event. Figure 12 shows a portion of the hidden state graph relating a target and query through a sequence of hidden states, where the dotted notes are examples of each generated note.

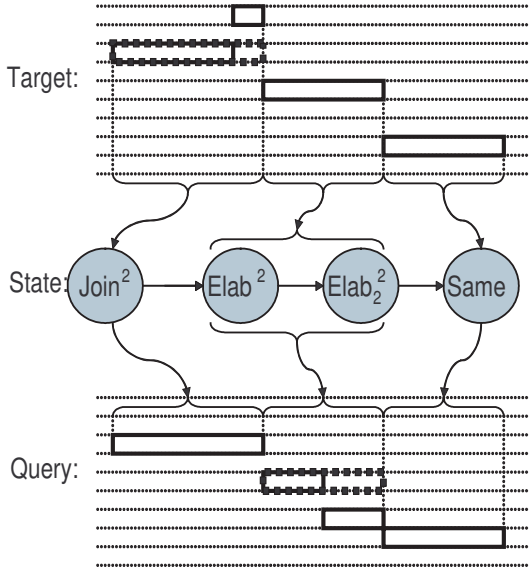

Figure 12: Relationship between states and events

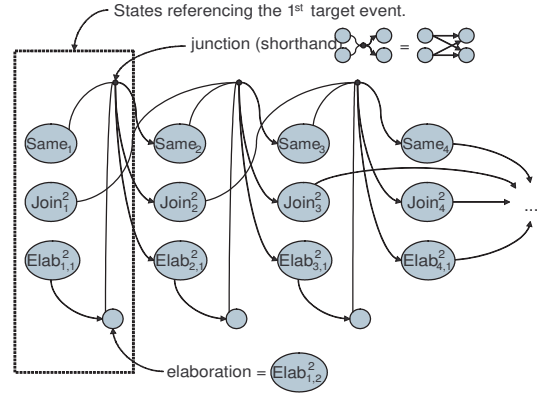

Figure 13: Edit type topology

Where $\langle Pitch[i], IOI[i] \rangle$ is the $i^{\text{th}}$ query note, and $\langle Pitch[t], IOI[t] \rangle$ the $t^{\text{th}}$ target note, we have the following expected relationships between target and query based on the hidden state at time $t$, $q_t = \langle E[t], 0, 0 \rangle$, ignoring transposition, tempo and local error for the

moment:

$$
\begin{cases}
\langle Pitch[i], IOI[i] \rangle = \langle Pitch[t], IOI[t] \rangle, & \text{if } E[t] = Same_i \\
\langle Pitch[i], \sum_{j=i}^{i+l-1} IOI[j] \rangle = \langle Pitch[t], IOI[t] \rangle, & \text{if } E[t] = Join_i^l \\
\langle Pitch[i], IOI[i] \rangle = \underbrace{\langle Pitch[t-j+1], \sum_{k=t}^{t+m-j} IOI[k] \rangle}_{\text{Notice that all events in the elaboration } point \text{ to a single larger event}} & \text{if } E[t] = Elab_{i,j}^m
\end{cases}
\tag{7}
$$

### 7.1.2 Transposition and Tempo

In order to account for the various ways in which target and query could be related, we must further refine our state definition to include *transposition* and *tempo* cluster. The intuition here is that the edit type determines the alignment of events between the query with the target (see Figure 12 for instance) and the cluster determines the exact relationship between those events.

Using pitch class, there are only twelve possible distinct transpositions, because of the modulus-12 relationship to pitch. While any offset will do, we set $\mathbf{K} = \{-5, -4, \dots, +6\}$. We establish limits on how far off target a query can be with respect to tempo, allowing the query to be between half- and double- speed. This corresponds to values in the range $\mathbf{S} = \{-4, -3, \dots, +4\}$ in terms of *quantized* tempo units (based on the logarithmic quantization scale described in Section B).

## 7.2 Transition Matrix

We now describe the transition matrix $A$, which maps from $\mathbf{S} \times \mathbf{S} \to \Re$. Where $q_t$ is the state at time $t$ (as defined by the position in the query, or observation sequence), $a_{xy}$ is equal to the probability $Pr(q_t = s_x | q_{t+1} = s_y, \lambda)$, or in other words, the probability of a transition from state $s_x$ to state $s_y$.

The transition probability is composed of three values, an edit type, modulation and tempo change probability:

$$
a_{xy} = a_{xy}^E \cdot a_{xy}^K \cdot a_{xy}^S
\tag{8}
$$

We describe each of these values individually.

### 7.2.1 Edit Type Transition

Most of the edit type transitions have zero probability, as suggested by the state descriptions. For instance, $Same_i$ states can only precede states pointing to index $i + 1$ in the target. Elaboration states are even more restrictive, as they form deterministic chains of the form: $Elab_{i,1}^m \to Elab_{i,2}^m, \to \dots \to Elab_{i,m}^m$. This last state can then proceed like $Same_i$, to the $i + 1$ states. Similarly, $Join_i^l$ states can only proceed to $i + l$ states. A sample model topology is shown in Figure 13, for $M = L = 2$. Note that this is a left-right model, in which transitions impose a partial ordering on states.

Based on properties of the target, we can generate these transition probabilities. We define $P_{Join}(i, l)$ as the probability that the $i^{\text{th}}$ note in the target will be modified by an order $l$ join. $P_{Elab}(i, m)$ is the probability that the $i^{\text{th}}$ note in the target will be modified by an order $m$ elaboration. $P_{Same}(i)$ has the expected meaning. Since every state has non-zero transitions to all states with a particular position in the target, we must insure that:

$$\forall i, P_{Same}(i) + \sum_{l=2}^{L} P_{Join}(i,l) + \sum_{m=2}^{M} P_{Elab}(i,m) = 1 \qquad (9)$$

This also implies that along non-zero transitions, the probability is entirely determined by the second state. For example, the probability of the transition $Join_3^2 \rightarrow Elab_{5,1}^2$ is the same as for $Same_4 \rightarrow Elab_{5,1}^2$.

Establishing separate distributions for every index in the target would be problematic. For this reason, we need to tie distributions by establishing equivalence classes for edit type transitions. Each equivalence class is a *context* for transitions, where the $k^{\text{th}}$ edit context is denoted $C_k^E$. A state that is a member of the $k^{\text{th}}$ edit context ($s_x \in C_k^E$) shares its transition probability function with all other members of that context. Each state $s_y$ has an associated $\Delta^{(E)}$ value, which is a classification according to the *type* (e.g. "Join") and *degree* (e.g. $l = 2$) of edit type. We define the function $P_k^E(\Delta^{(E)})$ as the probability of a transition to edit classification $\Delta^{(E)}$ in edit context $k$, so that for a transition $s_x \rightarrow s_y$:

$$a_{Xy}^E = P_k^E(\Delta^{(E)}) \leftrightarrow s_x \in C_k^E \text{ and } s_y \text{ has edit classification } \Delta^{(E)}. \qquad (10)$$

We intentionally leave the definition of context somewhat open. With reference to broadly observed trends in queries and their transcription, we suggest these alternatives:

- The simplest and easiest to train solution is simply to build up tables indicating the chances that, in general, a note will be elaborated or joined. Thus, the probabilities are independent of the particular event in the target. For instance, our current test implementation uses this approach with $M = 2$ and $L = 2$, with $P_{Same} = 0.95$, $P_{Join} = \{0.03\}$ and $P_{Elab} = \{0.02\}$.

- Transcribers, in our experience, are more likely to "miss" shorter notes, as are singers (consider for instance Figure 1, in which the second and third note are joined.) As such, we believe it will be possible to take advantage of contextual information (durations of surrounding events) to determine the likelihood of joins and elaborations at each point in the target sequence.

### 7.2.2 Modulation and Tempo Change

Modulation and tempo changes are modelled as transitions between clusters. We denote the probability of modulating by $\Delta^{(K)}$ semitones on the $i^{\text{th}}$ target event as $P_{Modu}(i, \Delta^{(K)})$ (again defined over the range $-5 \leq \Delta^{(K)} \leq +6$). The probability of a tempo change of $\Delta^{(S)}$ quantization units is denoted $P_{Change}(i, \Delta^{(S)})$, allowing for a halving to doubling of tempo at each step ($-4 \leq \Delta^{(S)} \leq +4$).

Again, we need to tie parameters by establishing contexts for transposition (denoted $C_i^K$ with associated probability function $P_i^K$) and tempo change (denoted $C_i^S$ with associated probability function $P_i^S$). Without restricting the definition of these contexts, we suggest the following alternatives, for modulation:

- In our current implementation, we simply apply a normal distribution over modulation centered at $\Delta^{(K)} = 0$, assuming that it is most likely a singer will not modulate on every note. The distribution is fixed across all events, so there is only one context.

- We may wish to take advantage of some additional musical context. For instance, we have noted that singers are more likely to modulate during a large pitch interval.

We have observed no clear trend in tempo changes. Again, we simply define a normal distribution centered at $\Delta^{(S)} = 0$.

### 7.2.3 Anatomy of a Transition

In a transition $s_x \rightarrow s_y$ (where $s_x = \langle E[x], K[x], S[x] \rangle$), $s_x$ belongs to three contexts: $C_i^E$, $C_j^K$ and $C_k^S$. The second state is an example of some edit classification $\Delta^{(E)}$, so $a_{xy}^E = P_i^E(\Delta^{(E)})$. The transition corresponds to a modulation of $\Delta^{(K)} = K[y] - K[x]$, so $a_{xy}^K = P_j^K(\Delta^{(K)})$. Finally, the transition contains a tempo change of $\Delta^{(S)} = S[y] - S[x]$, so $a_{xy}^S = P_k^S(\Delta^{(S)})$.

## 7.3 Initial State Distribution

We associate the initial state distribution in the hidden model with a single target event. As such, a separate model for each possible starting point must be built. Note, however, that we can actually reference a single larger model, and generate different initial state distributions for each separate starting-point model, addressing any concerns about the memory and time costs of building the models. Essentially, these various "derived" models correspond to various alignments of the start of the target with the query.

The probability of beginning in state $s_x$ is denoted $\pi_x$. As with transition probabilities, this function is composed of parts for edit type ($\pi_x^E$), transposition ($\pi_x^K$) and tempo ($\pi_x^S$).

Our initial edit distribution ($\pi_x^E$), for an alignment starting with the $i^{\text{th}}$ event in the target, is over only those edit types associated with $i$: $Same_i$, $\{Join_i^l\}_{l=2}^L$ and $\{Elab_{i,1}^m\}_{m=2}^M$. We tie initial edit probabilities to the edit transition probabilities, such that if $s_z$ directly precedes $s_x$ in the hidden-state topology, $\pi_x^E = a_{zx}^E$. This means that, for instance, the probability of a two-note join on the $i^{\text{th}}$ target event is the same whether or not $i$ happens to be the current starting alignment.

The initial distributions over transposition and tempo are as follows:

- $\pi_K(\chi)$: the probability of beginning a query in transposition $\chi$. Since the overwhelming majority of people do not have absolute pitch, we can make no assumption about initial transposition, and set $\pi_K(\chi) = \frac{1}{12}$, $-5 \leq \chi \leq +6$. This distribution could however be tailored to individual users' abilities, thus the distributions might be quite different between a musician with absolute pitch and a typical user.

- $\pi_S(\chi)$: the probability of beginning a query at tempo $\chi$. Since we are able to remember roughly how fast a song "goes", we currently apply a normal distribution[3] over initial tempo, with mean 0 and deviation $\sigma = 1.5$, again in the quantized tempo representation.

---

3. In our experiments, we frequently approximate normal distributions over a discrete domain, using the normal density function: $y = \frac{e^{\frac{-(\mu-x)^2}{2\sigma^2}}}{\sigma\sqrt{\pi}}$ , and then normalize to sum 1 over the function range.

## 7.4 Emission Function

Conventionally, a hidden state is said to emit an *observation*, from some discrete or continuous domain. A matrix $B$ maps from $\mathbf{S} \times \mathbf{O} \rightarrow \Re$, where $\mathbf{S}$ is the set of all states, and $\mathbf{O}$ is the set of all observations. $b_x(o_t)$ is the probability of emitting an observation $o_t$ in state $s_x$ ($Pr(o_t | q_t = s_x, \lambda)$). In our model, it is simpler to view a hidden state as emitting observation *errors*, relative to our expectation about what the pitch class and IOI should be based on the edit type, transposition and tempo.

Equation 7 defines our expectation about the relationship between target and query events given edit type. For the hidden state $s_x = \langle E[x], K[x], S[x] \rangle$, we will represent this relationship using the shorthand $\langle P[i], R[i] \rangle \rightarrow \langle P[t], R[t] \rangle$, mindful of the modifications suggested by the edit type. The pitch error is relative to the current transposition:

$$\Delta^{(P)} = P[t] - (P[i] + K[x]) \tag{11}$$

Similarly, we define an IOI error relative to tempo:

$$\Delta^{(R)} = R[t] - (R[i] + S[x]) \tag{12}$$

To simplify the parameterization of our model, we assume that pitch and IOI error are conditionally independent given state. For this reason, we define two emission probability functions, for pitch ($b_{s_x}^P(o_t)$) and rhythm ($b_x^R(o_t)$), where $b_x(o_t) = b_x^P(o_t) \cdot b_x^R(o_t)$. To avoid the need for individual functions for each state, we again establish equivalence classes, such that if $s_x \in C_i^P$, then $b_x^P(o_t) = P_i^P(\Delta^{(P)})$, using the above definition of $\Delta^{(P)}$. Similarly, $s_x \in C_i^R$ implies that $b_x^R(o_t) = P_i^R(\Delta^{(R)})$. This means that as a fundamental feature, we tie emission probabilities based on the error, reflecting the "meaning" of our states.

## 7.5 Alternative View

For expository purposes, we define state as a tuple incorporating edit, transposition and tempo information. Before proceeding, we will introduce an alternate view of state, which is useful in explaining the dependency structure of our model. In Figure 14.A, the first interpretation is shown. In the hidden states ($S$), each state is defined by $s_i = \langle E[i], K[i], S[i] \rangle$, and according to the first-order Markov assumption, the current state depends only on the previous state. Observations ($O$) are assumed to depend only on the hidden state, and are defined by $o_t = \langle P[t], R[t] \rangle$.

The second view provides more detail (Figure 14.B). Dependencies among the individual components are shown. The $E$, $K$ and $S'$ hidden chains denote the respective components of a hidden state. The edit type ($E$) depends only on the previous edit type (for a detailed illustration of this component, see Figure 13). The transposition ($K$) depends on both the previous transposition and the current edit type, since the degree of modulation and the current position in the target influence the probability of arriving at some transposition level. A pitch observation ($P$) depends only on the current edit type and the current transposition, which tell us which pitch we expect to observe: the "emission" probability is then simply the probability of the resulting error, or discrepancy between what we expect and what we see. There is a similar relationship between the edit type ($E$), tempo ($S'$), and rhythm observation ($R$).

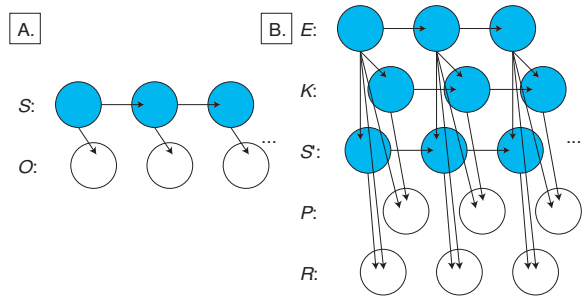

Figure 14: The dependencies in two views of the error model, where shaded circles are hidden states (corresponding to the target) and white circles are fully observed (corresponding to the query).

## 8. Probability of a Query

In the context of music retrieval, a critical task is the calculation of the likelihood that a certain target would generate a query given the model. Using these likelihood values, we can rank a series of potential database targets in terms of their relevance to the query.

Conceptually, the idea is to consider every possible path through the hidden model. Each path is represented by a sequence of hidden states $Q = \{q_1, q_2, \ldots, q_T\}$. This path has a probability equal to the product of the transition probabilities of each successive pair of states. In addition, there is a certain probability that each path will generate the observation sequence $O = \{o_1, o_2, \ldots, o_T\}$ (or, the query.) Thus, the probability of a query given the model (denoted $\lambda$) is:

$$Pr(O|\lambda) = \sum_{\forall Q} Pr(O|Q, \lambda) Pr(Q|\lambda) \tag{13}$$

$$= \sum_{\forall Q} \left[ \prod_{t=1}^{T} b_{q_t}(o_t) \right] \left[ \pi_{q_1} \prod_{t=2}^{T} a_{q_{t-1} q_t} \right] \tag{14}$$

Fortunately, there is considerable redundancy in the naïve computation of this value. The "standard" forward-variable algorithm (Rabiner, 1989) provides a significant reduction in complexity. This is a dynamic programming approach, where we inductively calculate the likelihood of successively longer suffixes of the query with respect to the model. We define a forward variable as follows:

$$\alpha_t(x) = Pr(\{o_1, o_2, \ldots, o_t\}, q_t = s_x | \lambda) \tag{15}$$

This is the probability of being in state $s_x$ at time $t$ given all prior observations. We initialize the forward variable using the initial state probabilities, and the observation probabilities over the initial observation:

$$\alpha_1(x) = Pr(\{o_1\}, q_t = s_x | \lambda) = \pi_x b_x(o_1) \tag{16}$$

By induction, we can then calculate successive values, based on the probabilities of the states in the previous time step:

$$\alpha_{t+1}(y) = \sum_{x=1}^{n} \alpha_t(x) a_{xy} b_y(o_{t+1}) \tag{17}$$

Finally, the total probability of the model generating the query is the sum of the probabilities of *ending* in each state (where $T$ is the total sequence length):

$$Pr(O|\lambda) = \sum_{x=1}^{n} \alpha_T(x) \tag{18}$$

## 8.1 Complexity Analysis

Based on the topology of the hidden model, we can calculate the complexity of the forward-variable algorithm for this implementation. Since each edit type has non-zero transition probabilities for at most $L + M - 1$ other edit types, this defines a branching factor ($b$) for the forward algorithm. In addition, any model can have at most $b|D|$ states, where $|D|$ is the length of the target.

Updating the transposition and tempo probabilities between two edit types (including all cluster permutations) requires $k = (9 \cdot 12)^2$ multiplications given the current tempo quantization, and the limits on tempo change. Notice that increasing either the allowable range for tempo fluctuation, or the resolution of the quantization, results in a super-linear increase in time requirements!

So, at each induction step (for $t=1, 2, \ldots$), we require at most $k|D|b^2$ multiplications. As such, for query length $T$, the cost is $O(k|D|b^2T)$. Clearly, controlling the branching factor (by limiting the degree of join and elaboration) is critical. $k$ is a non-trivial scaling factor, so we recommend minimizing the number of quantization levels as far as possible without overly sacrificing retrieval performance.

## 8.2 Optimizations

While asymptotic improvements in complexity are not possible, certain optimizations have proven quite effective, providing over a ten-fold improvement in running times. An alternate approach to calculating the probability of a query given the model is to find the probability of the most likely (single) path through the model, using the Viterbi algorithm. This is a classical dynamic programming approach, which relies on the observation that the optimal path must consist of optimal sub-paths. It works by finding the highest probability path to every state at time $t + 1$ based on the highest probability path to every state at time $t$. The algorithm is therefore a simple modification of the forward-variable algorithm. Instead of taking the sum probability of all paths leading into a state, we simply take the maximum probability:

$$\alpha_{t+1}(y) = \max_{x=1}^{n} [\alpha_t(x) a_{xy} b_y(o_{t+1})] \tag{19}$$

A side-effect of this change is that all arithmetic operations are multiplications for Viterbi (no summations.) As a result, we can affect a large speed-up by switching to a log-space, and adding log probabilities rather than multiplying.

Some other implementation details:

- The edit topology is quite sparse (see Figure 13), so it is advantageous to identify successors for each edit state rather than exhaustively try transitions.

- There is considerable redundancy in the feed-forward step (for both Viterbi and the forward-algorithm) since many state transitions share work. For instance, all transitions of the form $\langle E[x], K[x], \cup \rangle \rightarrow \langle E[y], K[y], \cup \rangle$ share several components: the same edit transition probability, the same modulation probability and the same pitch observation probability. By caching the product of those probabilities, we avoid both repeated look-ups and repeated multiplications or additions, a non-trivial effect when the depth of the nesting is considered over edit type, transposition and tempo.

### 8.2.1 Branch and Bound

Using Viterbi, it is possible to use branch and bound to preemptively prune paths when it can be shown that no possible completion can result in a high enough probability. First, we should explain what we mean by "high enough": if only a fixed number $(k)$ of results are required, we reject paths not capable of generating a probability greater or equal to the $k^{\text{th}}$ highest probability observed thus far in the database. How can we determine an upper-bound on the probability of a path? We note that each event (or observation) in an optimal Viterbi path introduces a factor, which is the product of the observation probability, the edit transition probability, and the inter-cluster transition probabilities. Knowing the maximum possible value of this factor $(f)$ allows us to predict the minimum "cost" of completing the algorithm along a given path. For instance, given a query of length $T$, and an interim probability of $\alpha_t(x)$, we can guarantee that no possible sequence of observations *along this path* can result in a probability greater than $\alpha_t(x) f^{T-t}$ We use this last value as a heuristic estimate of the eventual probability.

We can determine $f$ in several ways. Clearly, there is an advantage to minimizing this factor, though setting $f = 1$ is feasible (since no parameter of the model can be greater than 1). A simple and preferable alternative is to choose $f$ as the product of the maximum transition and maximum emission probabilities:

$$f = \max_{x,y=1}^{n} a_{xy} \cdot \max_{y=1}^{n} \left( \max_{\forall o} b_y(o) \right) \tag{20}$$

In effect, we are defining the behavior of the ideal query – a model parameterization in which there is no error, modulation or tempo change.

## 9. Training

We need to learn the following parameters for our HMM:

- the probabilities of observing all pitch and rhythm errors (the functions $P_c^P$ and $P_c^R$ for all contexts $c$);

- the probabilities of modulating and changing tempo by all relevant amounts ($P_c^K$ and $P_c^S$); and,

- the probabilities of transitions to each of the edit types ($P_c^E$).

We fix some parameters in our model. For instance, the initial edit type distributions are not explicitly trained, since as described, these are tied to the edit type transition function. In addition, we assume a uniform distribution over initial transposition and a normal distribution over initial tempo. This is because we see no way of generalizing initial distribution data to songs for which we have no training examples. Consider that, for instance, the tendency for users to sing "Hey Jude" sharp and fast should not be seen to influence their choice of transposition or tempo in "Moon River".

We will describe the training procedure in terms of a simple HMM, and then describe the extensions required for our model.

### 9.1 Training a Simple HMM

With a fully-observable Markov Model, it is fairly straightforward to learn transition probabilities: we simply count the number of transitions between each pair of states. While we cannot directly count transitions in an HMM, we can use the forward variable and a backward variable (defined below) to calculate our expectation that each hidden transition occurred, and thus "count" the number of transitions between each pair of states indirectly. Until we have parameters for the HMM, we cannot calculate the forward- and backward-variables. Thus we pick starting parameters either randomly or based on prior expectations, and iteratively re-estimate model parameters. This procedure is known as the Baum-Welch, or expectation-maximization algorithm (Baum & Eagon, 1970).

Consider a simple HMM (denoted $\lambda$) with a transition matrix $A$, where $a_{xy}$ is the probability of the transition from state $s_x$ to state $s_y$, an observation matrix $B$ where $b_y(o_t)$ is the probability of state $s_y$ emitting observation $o_t$, and an initial state distribution $\Pi$ where $\pi_x$ is the probability of beginning in state $s_x$. Given an observation sequence $O = \{o_1, o_2, \ldots, o_T\}$, we again define a forward variable, calculated according to the procedure defined in Section 8:

$$\alpha_t(x) = Pr(\{o_1, o_2, \ldots, o_t\}, q_t = s_x | \lambda) \tag{21}$$

In addition, we define a backward variable, the probability of being in a state given all subsequent observations:

$$\beta_t(x) = Pr(\{o_{t+1}, o_{t+2}, \ldots, o_T\}, q_t = s_x | \lambda) \tag{22}$$

We calculate values for the backward-variable inductively, as with the forward-variable, except working back from the final time step $T$:

$$\beta_T(x) = 1, \text{ arbitrarily} \tag{23}$$

$$\beta_{t-1}(x) = \sum_{y=1}^{n} a_{xy} b_y(o_t) \beta_t(y) \tag{24}$$

We define an interim variable $\xi_t(x, y)$, the probability of being in state $s_x$ at time $t$ and state $s_y$ at time $t+1$, given all observations:

$$\xi_t(x, y) = Pr(q_t = s_x, q_{t+1} = s_y | O, \lambda) \tag{25}$$

$$= \frac{Pr(q_t = s_x, q_{t+1} = s_y, O | \lambda)}{Pr(O | \lambda)} \tag{26}$$

$$= \frac{\alpha_t(x) a_{xy} b_y(o_{t+1}) \beta_{t+1}(y)}{\sum_{x=1}^{n} \sum_{y=1}^{n} \alpha_t(x) a_{xy} b_y(o_{t+1}) \beta_{t+1}(y)} \tag{27}$$

Finally, we introduce the variable $\gamma_t(x)$, the probability of being in state $s_x$ at time $t$. This can be derived from $\xi_t(x, y)$:

$$\gamma_t(x) = \sum_{y=1}^{n} \xi_t(x, y) \tag{28}$$

These values can be used to determine the *expected* probability of transitions and the expected probability of observations in each state, and thus can be used to re-estimate model parameters. Where the new parameters are denoted $\hat{\Pi}$, $\hat{A}$ and $\hat{B}$, we have:

$$\hat{\pi}_x = \gamma_1(x) \tag{29}$$

$$\hat{a}_{xy} = \frac{\overbrace{\sum_{t=1}^{T-1} \xi_t(x, y)}^{\text{expected number of transitions from } s_x \to s_y}}{\underbrace{\sum_{t=1}^{T-1} \gamma_t(x)}_{\text{expected number of transitions from } s_x}} \tag{30}$$

$$\hat{b}_y(o) = \frac{\overbrace{\sum_{t=1}^{T} \begin{cases} \gamma_t(y), & \text{if } o_t = o \\ 0 & \text{otherwise} \end{cases}}^{\text{expected number of times in state } s_y \text{ observing } o}}{\underbrace{\sum_{t=1}^{T} \gamma_t(y)}_{\text{expected number of times in state } s_y}} \tag{31}$$

By iteratively re-estimating the parameter values, we converge to a local maximum (with respect to the expectation of a training example) in the parameter space. In practice, the procedure stops when the parameter values change by less than some arbitrary amount between iterations.

## 9.2 Training the Query Error Model

Our query model has a few key differences to the model outlined above: heavy parameter tying, and multiple components for both transitions and observations. The procedure is fundamentally the same, however. Instead of asking "How likely is a transition from $s_x \rightarrow s_y$ (or what is $\hat{a}_{xy}$)?", we ask, for instance "How likely is a modulation of $\Delta^{(K)}$ in modulation context $c$ (or what is $\hat{P}_c^K(\Delta^{(K)})$)?" To answer this question, we define an interim variable,

$$\xi_t^K(\Delta^{(K)}, c) = \sum_{s_x \in C_c^K} \begin{cases} \xi_t(x, y), & \text{if } K[y] - K[x] = \Delta^{(K)} \\ 0 & \text{otherwise} \end{cases}, \tag{32}$$

the probability of a modulation of $\Delta^{(K)}$ in modulation context $c$ between time steps $t$ and $t+1$. We can now answer the question as follows:

$$\hat{P}_c^K(\Delta^{(K)}) = \frac{\sum_{t=1}^{T-1} \xi_t^K(\Delta^{(K)}, c)}{\sum_{t=1}^{T-1} \sum_{\chi=-5}^{6} \xi_t^K(\chi, c)} \tag{33}$$

We use a similar derivation for the other two components of a transition. For the edit type function, we have:

$$\xi_t^E(\Delta^{(E)}, c) = \sum_{s_x \in C_c^E} \begin{cases} \xi_t(xy), & \text{if } E[y] \text{ is an instance of } \Delta^{(E)} \\ 0 & \text{otherwise} \end{cases} \tag{34}$$

$$\hat{P}_c^E(\Delta^{(E)}) = \frac{\sum_{t=1}^{T-1} \xi_t^E(\Delta^{(E)}, c)}{\sum_{t=1}^{T-1} \sum_{\forall \Delta^{(E)'}} \xi_t^E(\Delta^{(E)'}, c)}; \tag{35}$$

and, for the tempo change function, we have:

$$\xi_t^S(\Delta^{(S)}, c) = \sum_{s_x \in C_c^S} \begin{cases} \xi_t(x, y), & \text{if } S[y] - S[x] = \Delta^{(S)} \\ 0 & \text{otherwise} \end{cases} \tag{36}$$

$$\hat{P}_c^S(\Delta^{(S)}) = \frac{\sum_{t=1}^{T-1} \xi_t^S(\Delta^{(S)}, c)}{\sum_{t=1}^{T-1} \sum_{\chi=-4}^{4} \xi_t^S(\chi, c)} \tag{37}$$

The emission function re-estimation is more straightforward. For pitch error, we have:

$$\hat{P}_c^P(\Delta^{(P)}) = \frac{\sum_{t=1}^{T} \sum_{s_y \in C_c^P} \begin{cases} \gamma_t(y), & \text{if observing } \Delta^{(P)} \text{ in this state} \\ 0 & \text{otherwise} \end{cases}}{\sum_{t=1}^{T} \sum_{s_y \in C_c^P} \gamma_t(y)}; \tag{38}$$

and, for rhythm error we have:

$$\hat{P}_c^R(\Delta^{(R)}) = \frac{\sum_{t=1}^{T} \sum_{s_y \in C_c^R} \begin{cases} \gamma_t(y), & \text{if observing } \Delta^{(R)} \text{ in this state} \\ 0 & \text{otherwise} \end{cases}}{\sum_{t=1}^{T} \sum_{s_y \in C_c^R} \gamma_t(y)}. \tag{39}$$

Again, we are simply "counting" the number of occurrences of each transition type and observation error, with the additional feature that many transitions are considered evidence for a particular context, and every transition is in turn considered evidence for several contexts. A formal derivation of the re-estimation formulae is given in Appendix A.

### 9.3 Starting Parameters

The components of our model have clear musical meanings, which provide guidance for the selection of starting parameters in the training process. We apply normal distributions over the error and cluster change parameters, centered about "no error" and "no change", respectively. This is based solely on the conjecture (without which the entire MIR exercise would be a lost cause) that singers are in general more likely to introduce small errors than large ones. Initial edit probabilities can be determined by the hand-labelling of a few automatically transcribed queries. It is important to make a good guess at initial parameters, because the re-estimation approach only converges to a *local* maximum.

## 10. Analysis

To maintain generality in our discussion, and draw conclusions not specific to our experimental data or approach to note representation, it is useful to analyze model entropy with respect to cumulative and local error. What influences retrieval performance? From an information perspective, entropy provides a clue. Intuitively, the entropy measures our uncertainty about what will happen next in the query. Formally, the entropy value of a process is the mean amount of information required to predict its outcome. When the entropy is higher, we will cast a wider net in retrieval, because our ability to anticipate how the singer will err is reduced.

What happens if we assume cumulative error with respect to pitch when local error is in fact the usual case? Consider the following simplified analysis: assume that two notes are generated with pitch error distributed according to a normal Gaussian distribution, where $X$ is the random variable representing the error on the first note, and $Y$ represents the second. Therefore we have: $f_X(x) = Pr(X = x) = \frac{1}{\sqrt{2\pi}}e^{\frac{-x^2}{2}}$ and $f_Y(y) = Pr(Y = y) = \frac{1}{\sqrt{2\pi}}e^{\frac{-y^2}{2}}$. What is the distribution over the error on the *interval*? If $Z$ is the random variable representing the interval error, we have: $Z = Y - X$. Since $f_X(x)$ is symmetrical about $x = 0$, where $*$ is the convolution operator, we have: $f_Z(z) = f_X * f_Y(z) = \frac{1}{\sqrt{4\pi}}e^{\frac{-z^2}{4}}$, which corresponds to a Gaussian distribution with variance $\sigma^2 = 2$ (as compared with a variance of $\sigma^2 = 1$ for the local error distribution). Given this analysis, the derivative entropy for local error is equal to $\frac{1}{2}(\log(2\pi\sigma^2) + 1) \approx 1.42$, and the derivative entropy of the corresponding cumulative error is roughly 1.77. The underlying distributions are shown in Figure 15. It is a natural intuition that when we account for local error using cumulative error – as is implicitly done with intervallic pitch representations – we flatten the error distribution.

While experimental results indicate that local error is most common, sweeping cumulative error under the rug can also be dangerous, particularly with longer queries. When we use local error to account for a sequence of normally distributed cumulative errors represented by the random variables $X_1, X_2, \ldots, X_n$, the local error ($Z$) must absorb the *sum over all previous cumulative errors*: $Z = \sum_{i=1}^{n} X_i$. For example, when a user sings four consecutive notes cumulatively sharp one semi-tone, the final note will be, in the local view, four semi-tones sharp. If cumulative error is normally distributed with variance $\sigma^2$, the expected distribution on local error after $n$ notes is normally distributed with variance $n\sigma^2$ (a standard result for the summation of Gaussian variables). As such, even a low probability of cumulative error can hurt the performance of a purely local model over longer queries.

The critical observation here is that each simplifying assumption results in the compounding of error. Unless the underlying error probability distribution corresponds to an impulse function (implying that no error is expected), the summation of random variables always results in an increase of entropy. Thus, we can view these results as fundamental to any retrieval mechanism.

## 11. Results

160 queries were collected from five people – who will be described as subjects A-E, none involved in MIR research. Subject A is a professional instrumental musician, and subject C has some pre-college musical training, but the remaining subjects have no formal musical background. Each subject was asked to sing eight passages from well-known songs. We recorded four versions of each passage for each subject, twice with reference only to the lyrics of the passage. After these first two attempts, the subjects were allowed to listen to a MIDI playback of that passage – transposed to their vocal range – as many times as needed to familiarize themselves with the tune, and sang the queries two more times.

### 11.1 Training

JCS can be configured to support only certain kinds of error by controlling the starting parameters for training. For instance, if we set the probability of a transposition to zero, the re-estimation procedure will maintain this value throughout.

The results of this training, for three versions of the model over the full set of 160 queries, are shown in Figure 16, which indicates the overall parameters learned for each model. For all versions, a similar edit distribution results: the probability of no edit is roughly 0.85, the probability of consolidation is 0.05 and the probability of fragmentation is 0.1. These values are related primarily to the behavior of the underlying note segmentation mechanism.

In one of the models, both local and cumulative error are considered, labelled "Full" in the figure. Constrained versions, with the expected assumptions, are labelled "Local" and "Cumulative" respectively. It should be apparent that the full model permits a tighter distribution over local error (rhythm error and pitch error) than the simplified local model, and a tighter distribution over cumulative error (tempo change and modulation) than the simplified cumulative model.

When JCS has the luxury of considering both cumulative and local error, it converges to a state where cumulative error is nonetheless extremely unlikely (with probability 0.94 there is no change in tempo at each state, and with probability of 0.93 there is no modulation), further evidence (Pollastri, 2001) that local error is indeed the critical component. This flexibility however allows us to improve our ability to predict the local errors produced by singers, as evidenced by the sharper distribution as compared with the purely local version. The practical result is that the full model is able to *explain the queries in terms of the fewest errors*, and converges to a state where the queries have the highest expectation.

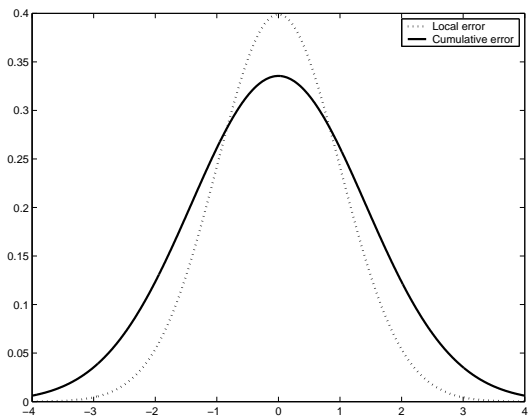

Figure 15: Assuming cumulative error when error is local

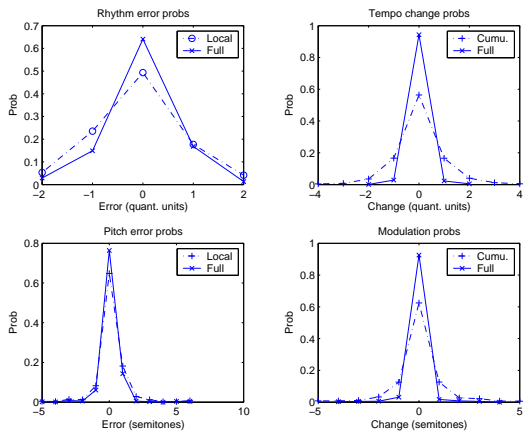

Figure 16: Result of training

## 11.2 Retrieval Performance

Given the analysis in Section 10, it is interesting to consider the effects on retrieval performance when we assume that only local or only cumulative error occurs. To this end, we generated a collection of 10000 synthetic database songs, based on the statistical properties (pitch intervals and rhythmic relationships) of a 300 piece collection of MIDI representations of popular and classical works. In our experiments, we compare several versions of JCS:

1. 'Full' model: this version of JCS models both local and cumulative error.

2. 'Restricted' model: a version of the full model which limits the range of tempo changes and modulations ($\pm 40\%$ and $\pm 1$ semitone respectively). This seems like a reasonable approach because training reveals that larger cumulative errors are extremely infrequent.

3. 'Local' model: only local error is modelled.

4. 'Cumulative' model: only cumulative error is modelled.

We first randomly divided our queries into two sets for training the models and testing respectively. After training each of the models on the 80 training queries, we evaluated retrieval performance on the remaining 80 testing queries. In evaluating performance, we consider the rank of the correct target's match score, where the score is determined by the probability that each database song would "generate" the query given our error model. In case of ties in the match score, we measure the *worst-case* rank: the correct song is counted below all songs with an equivalent score. In addition to the median and mean rank, we provide the mean reciprocal rank (MRR): this is a metric used by Text REtrieval Conference (Voorhees & Harman, 1997) to measure text retrieval performance. If the ranks of the correct song for each query in a test set are $r_1, r_2, \ldots, r_n$, the MRR is equal to, as the name suggests: $\frac{1}{n} \sum_{i=1}^{n} \frac{1}{r_i}$.

| | Full | Restricted | Local | Cumulative | Simple HMM |
|---|---|---|---|---|---|
| MRR | 0.7778 | 0.7602 | 0.7483 | 0.3093 | 0.4450 |
| Median | 1 | 1 | 1 | 68.5 | 5 |
| Mean | 490.6 | 422.9 | 379.5 | 1861 | 1537.5 |

Table 1: Overview of rank statistics

The distribution of ranks is summarized in Figure 18. The rank statistics are shown in Table 1. In addition, we provide the Receiver Operating Characteristic (ROC) curves for each model (see Figure 17) which indicate the precise tradeoffs between sensitivity and specificity, as characterized by the true positive and false positive rates for the model. Because there is only one true example in each test, we are concerned primarily with results where the false positive rate is very low. For this reason, we present false positives on a logarithmic scale. Note that a false positive rate of $10^{-4}$ corresponds to a single false positive for the size database we are using. For reference, we indicate the performance of a theoretical random model. For comparison, we provide results from another HMM-based QBH system, listed as "Simple HMM" (Shifrin et al., 2002), using default parameters. This system uses a relative note representation, and has hidden states corresponding to quantized instances of the pitch interval and IOI ratio pairs found in the target.

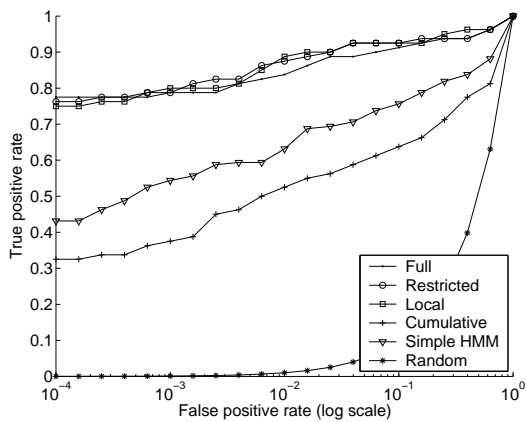

Figure 17: ROC curves for various JCS implementations *false positive rate shown on log scale

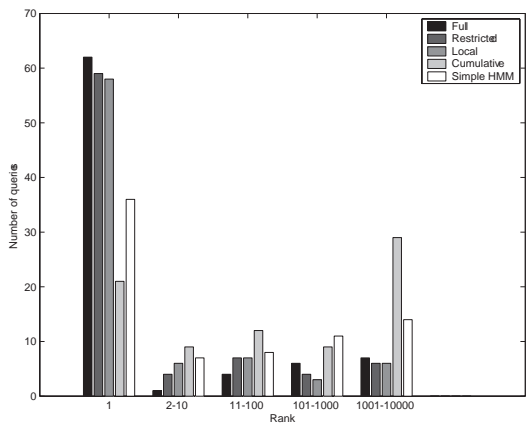

Figure 18: Distribution of ranks

The cumulative error model and the simple model perform quite poorly in comparison with the other approaches, owing to the prevalence of local error in our query collection. We see little evidence of the reverse phenomenon: notice that restricting or ignoring cumulative error does not have a notable impact on retrieval performance except on the longest queries, where there is evidence of degradation in performance using the local approach. This result agrees with the basic entropy analysis, which predicts greater difficulty for 'local' approaches on longer queries.

|        | Full          | Restricted    | Local         | Cumulative    |
| ------ | ------------- | ------------- | ------------- | ------------- |
| MRR    | 0.8011/0.8911 | 0.8320/0.8949 | 0.8345/0.8676 | 0.2669/0.6047 |
| Median | 1/1           | 1/1           | 1/1           | 27/1          |
| Mean   | 37/17.9       | 33.6/21.6     | 43.7/21.5     | 150.8/79.0    |

Table 2: Results overview for generalization experiments, with baseline results given (*result/baseline*)

It is informative to examine where JCS fails. We identify two classes of failure:

- Alignment assumption failure: This is the most common type of error. JCS assumes that the entire query is contained in the database. When the segmenter misclassifies regions before and after the query proper as notes, this situation arises. JCS must explain the *entire* query in the context of each target, including these margins. JCS does however model such added notes *within* the query, using the elaboration operation.

- Entropy failure: errors are so prevalent in the query that many target to query mappings appear equally strong. Interestingly, we achieve solid performance in many cases where the queries are – subjectively – pretty wildly off the mark. While using a different underlying representation might allow us to extract additional useful information from queries, this does not alter the fundamental conclusions drawn about retrieval behavior with different approaches to error.

### 11.3 Training Generalization

Because of the redundancy in the query collection – multiple versions of each of the passages, and multiple examples of each subject's singing – it is informative to examine retrieval performance when the models are trained using examples completely unrelated to the test set. We randomly selected four of the eight passages in our study, and two of the five subjects, and used these queries for training. The remaining passages, performed by the remaining subjects, were used for training. In this way, the two sets do not share either singers or passages.

As a baseline, we provide retrieval results when the model is trained on the test set (see Table 2, Figures 19 and 20).

In these experiments, we see evidence of over-fitting for the more heavily parameterized "full" model, attributable in some part to the relatively small training set available for this experiment. There is a considerable difference in performance between the baseline and generalization results. The restricted model is a good compromise, because the relatively small number of parameters limits the risk of over-fitting, while modelling a broad range of errors.

### 11.4 Run-time Performance

While various parametrization-specific optimizations are possible, the general implementation of JCS has lowest-common-denominator behavior with respect to running time performance. Running on a 1.6GHz Pentium 4 Linux machine, a Java version of JCS performs on average 15 query to target comparisons per second. Training converges to a local-maximum after 16 iterations on average, or just over seven minutes on the same system using 80 training examples.

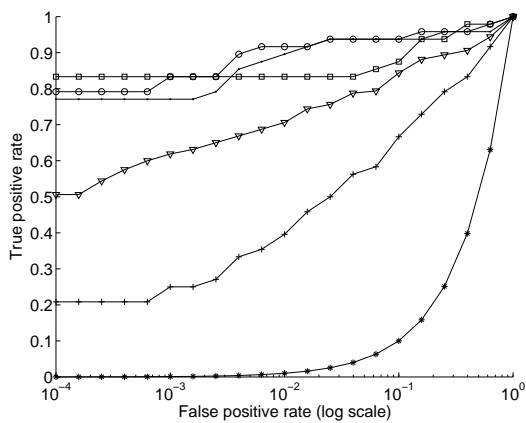

Figure 19: Retrieval performance for un-related test and training sets

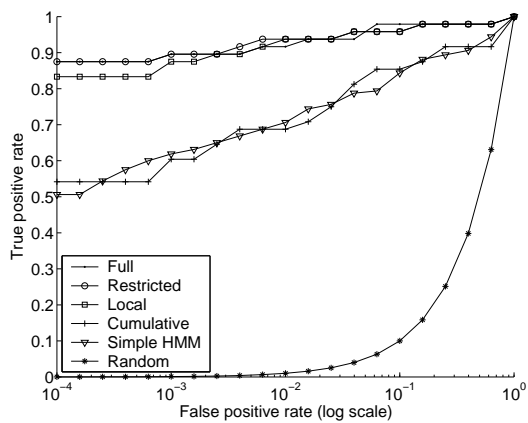

Figure 20: Baseline performance

## 12. Future Work

Even with the generalizations described in this model, a large number of parameters remain. We are currently gathering query data to train the model, as more in-depth evaluations of performance on non-synthetic queries will be essential. Various important questions remain to be answered, such as the following:

- What is the effect of query representation, for instance using a conventional note representation rather than pitch-class?

- How can we best tie parameters for training? For efficient training, how many contexts can (or should) be established?

- HMMs are amenable to "frame-based" representations, which would allow us to bypass the problematic note-segmentation stage of query transcription. Instead of modelling the query as a sequence of discrete note events, it is represented as a sequence of fixed-width time-frame analyses. Each state in the target model then has an associated distribution over duration - the probability of remaining in the state for some number of time-frames. We would like to explore the effectiveness of this approach, particularly with regards to the tradeoffs between time and retrieval performance.

Finally, tests on much larger databases will be necessary. While we believe that meta-data in the query process (genre, era, instrumentation) will allow us to restrict searches to a subset of a database or library, it is reasonable to assume that a large number of targets will be relevant to many searches.

## 13. Conclusion

We have presented a comprehensive model for singer error when creating a musical query. Some of these errors are tempo, pitch (both octave and absolute), melodic contour, and the particularly difficult class of joins ("skipped" notes in the query) and elaborations (note inserted into the query). It has been shown that a natural "musical" interpretation of these edits helps alleviate the problem of target similarity when edits are observed, noted in an earlier QBH feasibility study (Sorsa, 2001).

Our experimental results show that the model is effective in handling these errors. By this, we mean that we can identify when these errors occur and can account for them in the retrieval task when necessary. The experimental results indicate good retrieval performance when these errors occur in queries. Moreover, our model is able to separately model cumulative error and local error, or a combination of both. Generally, music-information retrieval researchers have been divided on the importance of these errors. We have been able to both theoretically and experimentally show that local error is more important. That is, ignoring cumulative error, in general, has a less deleterious effect on retrieval performance than ignoring local error. In addition, modelling both errors is may give improved performance, but at a computation cost.

Finally, our model demonstrates the most parsimonious way to model errors. We use the idea of a distributed state representation in our HMM to express the difference between a query and a target. This allows us to keep the model compact, improving both runtime efficiency and retrieval precision.

## Acknowledgements

We gratefully acknowledge the support of the National Science Foundation under grant IIS-0085945, and The University of Michigan College of Engineering seed grant to the MusEn project. The opinions in this paper are solely those of the authors and do not necessarily reflect the opinions of the funding agencies.

## Appendix A. Deriving Re-estimation Formulae

The reestimation procedure converges to a critical point in the parameter space with respect to likelihood. Baum defines an *auxiliary* function $\mathbf{q}$, where $\lambda'$ represents the "current" model parameter values, and we are attempting to iteratively reestimate $\lambda$:

$$\mathbf{q}(\lambda', \lambda) = \sum_Q P(O, Q|\lambda') \log P(O, Q|\lambda) \tag{40}$$

By maximizing this function, we maximize $P(O|\lambda)$ because:

$$\mathbf{q}(\lambda', \lambda) \geq \mathbf{q}(\lambda', \lambda') \Rightarrow P(O|\lambda) \geq P(O|\lambda') \tag{41}$$

We will now derive this implication. Notice that $P(O|\lambda) = \sum_Q P(O, Q|\lambda)$, or the sum of the probabilities of all possible paths through the model. There are a finite number $N$ of paths. Where $Q_i$ is the $i^{\text{th}}$ path, $p_i = P(O, Q_i|\lambda')$, $\sum$ is shorthand for $\sum_{i=1}^{N}$ and $q_i = P(O, Q_i|\lambda)$, we can rewrite the implication:

$$\sum p_i \log q_i \geq \sum p_i \log p_i \Rightarrow \sum q_i \geq \sum p_i \tag{42}$$

The derivation is as follows:

$$\sum p_i \log q_i \geq \sum p_i \log p_i \tag{43}$$

$$\sum p_i \log \frac{q_i}{p_i} \geq 0 \tag{44}$$

Since $x - 1 \geq \log x$, we can deduce from Equation 44 the following inequality:

$$\sum p_i (\frac{q_i}{p_i} - 1) \geq \sum p_i \log \frac{q_i}{p_i} \geq 0 \tag{45}$$

$$\sum q_i - \sum p_i \geq \sum p_i \log q_i - \sum p_i \log p_i \tag{46}$$

We know from the original implicant that $\sum p_i \log q_i \geq \sum p_i \log p_i$, so the right-hand side of Equation 46 is known to be non-negative. Therefore, the left-hand side of the inequality must also be non-negative. Since $P(O|\lambda) = \sum q_i$ and $P(O|\lambda') = \sum p_i$, it is clear that $P(O|\lambda) - P(O|\lambda') \geq 0$ and therefore $P(O|\lambda) \geq P(O|\lambda')$.

To derive our reestimation formulae, we first decompose $\mathbf{q}(\lambda'|\lambda)$ into a sum of auxiliary functions of the form:

$$f(\mathbf{y}) = \sum_{j=1}^{N} w_j \log y_j, \tag{47}$$

Where we are constrained by $y_i \geq 0$ and $g(\mathbf{y}) = \sum_i^N y_i = 1$ (a discrete probability function), the auxiliary functions can then be individually maximized using Lagrange multipliers. In general, where $\nabla g(\mathbf{y})$ is the gradient of the function $g(\mathbf{y})$, we know that extremes values of the function $f(\mathbf{y})$ subject to constraint $g(\mathbf{y})$ are solutions to the equation: $\nabla g(\mathbf{y}) = k \nabla f(\mathbf{y})$ where $k$ is some constant:

$$\left( \nabla g(\mathbf{y}) = \begin{bmatrix} \frac{\partial g}{\partial y_1} = 1 \\ \frac{\partial g}{\partial y_2} = 1 \\ \vdots \\ \frac{\partial g}{\partial y_N} = 1 \end{bmatrix} \right) = \left( k \nabla f(\mathbf{y}) = \begin{bmatrix} \frac{kw_1}{y_1} \\ \frac{kw_2}{y_2} \\ \vdots \\ \frac{kw_N}{y_N} \end{bmatrix} \right) \tag{48}$$

$$1 = \frac{kw_i}{y_i} \tag{49}$$

$$y_j = \frac{w_j}{\sum_{1=1}^{N} w_i} \text{ from the constraint } \sum_{i=1}^{N} w_i = 1 \tag{50}$$

To reduce **q** to this form, we rewrite:

$$P(O, Q|\lambda) = \pi_{q_1} \prod_{t=2}^{T} a_{q_{t-1}q_t} b_{q_t}(o_t) \tag{51}$$

Incorporating the various components of transition and emission probabilities, we have:

$$P(O, Q|\lambda) = \pi_{q_1} \prod_{t=2}^{T} a_{q_{t-1}q_t}^{E} a_{q_{t-1}q_t}^{K} a_{q_{t-1}q_t}^{S} b_{q_t}^{P}(o_t) b_{q_t}^{R}(o_t) \tag{52}$$

Finally, we incorporate the notion of context $(i, j, k, l, m)$ and amount/error/symbol $(\Delta)$, determined according to the procedure defined in Section 7:

$$P(O, Q|\lambda) = P_i^E(\Delta^{(E)}) \prod_{t=2}^{T} P_i^E(\Delta^{(E)}) P_j^K(\Delta^{(K)}) P_k^S(\Delta^{(S)}) P_l^P(\Delta^{(P)}) P_m^R(\Delta^{(R)}) \tag{53}$$

Converting to a log-scale, we have:

$$\sum_{t=1}^{T} \log P_i^E(\Delta^{(E)}) + \sum_{t=2}^{T} \log P_j^K(\Delta^{(K)}) + \tag{54}$$

$$\sum_{t=2}^{T} \log P_k^S(\Delta^{(S)}) + \sum_{t=2}^{T} \log P_l^P(\Delta^{(P)}) + \sum_{t=2}^{T} \log P_m^R(\Delta^{(R)})$$

Using this derivation, we rewrite **q**:

$$\mathbf{q}(\lambda', \lambda) = \sum_{\forall i} \mathbf{q}_i^E(\lambda', \mathbf{a}_x^E) + \sum_{\forall j} \mathbf{q}_j^K(\lambda', \mathbf{a}_x^K) + \tag{55}$$

$$\sum_{\forall k} \mathbf{q}_k^S(\lambda', \mathbf{a}_x^S) + \sum_{\forall l} \mathbf{q}_l^P(\lambda', \mathbf{a}_x^P) + \sum_{\forall m} \mathbf{q}_m^R(\lambda', \mathbf{a}_x^R)$$

where

$$\mathbf{q}_i^E(\lambda', \mathbf{a}_i^E) = \sum_{t=1}^{T} \sum_{\forall \Delta^{(E)}} P(O, q_t \in C_i^E, q_t \text{ instance of } \Delta^{(E)}|\lambda') \log P_i^E(\Delta^{(E)}) \tag{56}$$

$$\mathbf{q}_j^K(\lambda', \mathbf{a}_j^K) = \sum_{t=2}^{T} \sum_{\forall \Delta^{(K)}} P(O, q_{t-1} \to q_t \text{ instance of } \Delta^{(K)}, q_t \in C_j^K|\lambda') \log P_j^K(\Delta^{(K)}) \tag{57}$$

$$\mathbf{q}_k^S(\lambda', \mathbf{a}_k^S) = \sum_{t=2}^{T} \sum_{\forall \Delta^{(S)}} P(O, q_{t-1} \to q_t \text{ instance of } \Delta^{(S)}, q_t \in C_k^S|\lambda') \log P_k^S(\Delta^{(S)}) \tag{58}$$

$$\mathbf{q}_l^P(\lambda', \mathbf{b}_l^P) = \sum_{t=1}^{T} P(O, q_t \to o_t \text{ instance of } \Delta^{(P)}, q_t \in C_l^P|\lambda') \log P_l^P(\Delta^{(P)}) \tag{59}$$

$$\mathbf{q}_m^R(\lambda', \mathbf{b}_m^R) = \sum_{t=1}^{T} P(O, q_t \to o_t \text{ instance of } \Delta^{(R)}, q_t \in C_m^R|\lambda') \log P_m^R(\Delta^{(R)}) \tag{60}$$

Using the result in Equation 50, it is then trivial to derive the reestimation equations described in Section 9.

## Appendix B. Notation

We will now outline the notation used in describing the error model:

| Notation | Description |
|---|---|
| $\langle Pitch[x], IOI[x]\rangle$ | $x^{\text{th}}$ note event |
| $\langle P[x], R[x]\rangle$ | $x^{\text{th}}$ note event, quantized |
| $o_t$ and $d_i$ | $t^{\text{th}}$ observation (query note) and $i^{\text{th}}$ target event (database note) respectively |
| $\Delta^{(P)} = -1$ | a pitch error, one semi-tone flat |
| $\Delta^{(R)} = +1$ | a rhythm error, one quantization unit too long |
| $s_x = \langle E[x], K[x], S[x]\rangle$ $= \langle Same_1, +2, -3\rangle$ | $x^{\text{th}}$ HMM hidden state $E[x] = Same_1$: *Edit* type, replacement of first target note $K[x] = +2$: transposition (*Key*), 2 semi-tones sharp $S[x] = -3$: tempo *Scaling*, 3 units faster |
| $a_{xy} = a_{xy}^E \cdot a_{xy}^K \cdot a_{xy}^S$ | probability of a transition from hidden state $s_x \to s_y$ |
| $a_{xy}^E = P_t^E(\Delta^{(E)})$ | edit transition probability, with edit symbol $\Delta^{(E)}$ in context $C_t^E$ |
| $a_{xy}^K = P_t^K(\Delta^{(K)})$ | probability of a modulation $\Delta^{(K)}$ in context $C_t^K$ |
| $a_{xy}^S = P_k^S(\Delta^{(S)})$ | probability of a tempo change $\Delta^{(S)}$ in context $C_k^S$ |
| $\Delta^{(E)} \in \{Same_i\} \cup \{Join_1^i\}_{i=2}^L$ $\cup \{Elab_{i,j}^m\}_{m=2,j=1}^{M,m}$ | set of all possible edit symbols, where $L$ and $M$ are the "order" of the edit topology |
| $\alpha_t(x) = P(\{o_1, o_2, \ldots, o_t\}, q_t = s_x \vert \lambda)$ | forward-variable, the probability of ending in state $t$ at time $t$ given model $\lambda$ |
| $\beta_t(x) = P(\{o_{t+1}, o_{t+2}, \ldots, o_T\}, q_t = s_x \vert \lambda)$ | backward-variable, the probability of beginning in state $x$ at time $t$ |